\documentclass[12pt,titlepage]{article}
\usepackage[letterpaper,top=1in,bottom=1in,left=1.25in,right=1.25in]{geometry}
\usepackage{graphicx,cite, url}
\usepackage{color}
\usepackage{threeparttable}
\usepackage{amsmath,amssymb}
\usepackage{makecell}
\usepackage{multirow}
\usepackage{float}
\usepackage[linesnumbered,ruled]{algorithm2e}
\newenvironment{subroutine}[1][htb]
  {
   \begin{algorithm}[#1]%
  }{\end{algorithm}}

\begin{document}
\title{Structured illumination microscopy with unknown patterns and a statistical prior}

\author{Li-Hao Yeh$^1$, Lei Tian$^{1, 2}$ and Laura Waller$^{1,*}$\\
\\
\multicolumn{1}{p{\textwidth}}{\centering\emph{\normalsize 1. Department of Electrical Engineering and Computer Sciences, University of California, Berkeley, 94720, USA\\
2. Electrical \& Computer Engineering, Boston University, Boston, MA 02215, USA\\
$^{*}$  lihao.yeh@berkeley.edu
}}}

\maketitle

\begin{abstract}
Structured illumination microscopy (SIM) improves resolution by down-modulating high-frequency information of an object to fit within the passband of the optical system. Generally, the reconstruction process requires prior knowledge of the illumination patterns, which implies a well-calibrated and aberration-free system. Here, we propose a new \textit{algorithmic self-calibration} strategy for SIM that does not need to know the exact patterns {\it a priori}, but only their covariance. The algorithm, termed PE-SIMS, includes a Pattern-Estimation (PE) step requiring the uniformity of the sum of the illumination patterns and a SIM reconstruction procedure using a Statistical prior (SIMS). Additionally, we perform a pixel reassignment process (SIMS-PR) to enhance the reconstruction quality. We achieve 2$\times$ better resolution than a conventional widefield microscope, while remaining insensitive to aberration-induced pattern distortion and robust against parameter tuning. 
\end{abstract}

\section{Introduction}

The Abbe diffraction limit was considered to be the fundamental limit for spatial resolution of an optical microscope for more than a hundred years. In the last decade, novel techniques have circumvented this limit in order to achieve super-resolution~\cite{Lukosz1967,Neil1997,Heintzmann1999,Gustafsson2000,betzig2006imaging,Rust2006,Hell1994}. Structured illumination microscopy (SIM)~\cite{Lukosz1967,Neil1997,Heintzmann1999,Gustafsson2000}, for example, uses illumination by multiple structured patterns to down-modulate high spatial frequency information of the object into the low-frequency region, which can then pass through the bandwidth of the microscope's optical transfer function (OTF) and be captured by the sensor. The reconstruction algorithm for SIM combines demodulation process which brings the high spatial frequency information back to its original position and synthetic aperture that extends the support of the effective OTF. Various structured patterns have been used to realize SIM: periodic gratings~\cite{Lukosz1967,Neil1997,Heintzmann1999,Gustafsson2000}, a single focal spot (confocal microscope)~\cite{Wilson1984,Colin2003}, multifocal spots~\cite{York2012,York2013,Strohl2015,Chakrova2015} and random speckles~\cite{Garcia2005, Sylman2010, Mudry2012, Min2013, Dong2014, Yilmaz2015, Kim2015, Chakrova2015, Negash2016, Labouesse2016}. When the illumination patterns themselves are diffraction-limited, linear SIM is restricted to $2\times$ the bandwidth of a widefield microscope~\cite{Gustafsson2000}, allowing up to $\sim 2.4\times$ resolution enhancement (metrics explained in Sec.~\ref{Sec: Def_of_res}). 

In practice, structured illumination systems are sensitive to aberrations and experimental errors. To avoid reconstruction artifacts that degrade resolution, the patterns that are projected onto the sample must be known accurately. Periodic grating patterns can be parameterized by their contrast, period and phase angle, which may be estimated in the post-processing~\cite{Shroff2009,Shroff2010,Wicker2013,Wicker2013-2}. For multifocal patterns, the location of each focal spot is required~\cite{York2012}. For random speckle patterns, the relative shifts of the patterns are needed~\cite{Dong2014,Yilmaz2015}. Even with careful calibration and high-quality optics, distortions caused by the sample may degrade the result.  

To alleviate some of the experimental challenges, blind SIM was proposed, enabling SIM reconstruction without many priors~\cite{Mudry2012, Ayuk2013, Min2013, Jost2015, Negash2016, Labouesse2016}. The only assumption is that the sum of all illumination patterns is uniform. Optimization-based algorithms have been adopted, including iterative least squares with positivity and equality constraints~\cite{Mudry2012, Ayuk2013, Negash2016}, joint support recovery~\cite{Min2013} and $\ell_1$ sparsity constraints~\cite{Labouesse2016}. However, these algorithms are sensitive to parameter tuning and may show low contrast in reconstructing high spatial frequencies~\cite{Mudry2012}. Another algorithm, speckle super-resolution optical fluctuation imaging (S-SOFI) realizes SOFI~\cite{Dertinger2009} by first projecting random speckle patterns onto the object, and then using the statistical properties of the speckle patterns as a prior to reconstruct a high-resolution image~\cite{Kim2015}. S-SOFI is experimentally simple and robust; however it only achieves a $1.6 \times$ resolution enhancement instead of $2.4\times$ for conventional SIM techniques (as compared to a widefield microscope). 

In this paper, we propose a new reconstruction algorithm for SIM that is applicable to any illumination patterns. Our method, termed pattern estimation structured illumination microscopy with a statistical prior (PE-SIMS), is as robust and insensitive to parameter tuning as S-SOFI, and achieves better resolution enhancement (up to $2\times$). Like blind SIM, the patterns need not be known (except for a requirement on the covariance of the patterns). We demonstrate our method using simulated and experimental results with both speckle and multifocal patterns. We discuss pattern design strategies to reduce the amount of data required and demonstrate an extension that uses pixel reassignment~\cite{Cox1982, Sheppard1988, Muller2010, Sheppard2013,Roth2013} to improve the reconstruction quality. 

\section{Theory and Method}

Our algorithm takes in a SIM dataset consisting of multiple images captured under different structured illumination patterns (e.g. random speckles, multifocal spots). We reconstruct the super-resolved image in two parts. The first part is an iterative optimization procedure for estimating each illumination pattern based on an approximated object. The second part reconstructs the high-resolution image using the estimated patterns and the measured images, along with a statistical prior. Before introducing these two parts, we start by defining the SIM forward model. 

\begin{figure}[h]
\centering
\includegraphics[width=13cm]{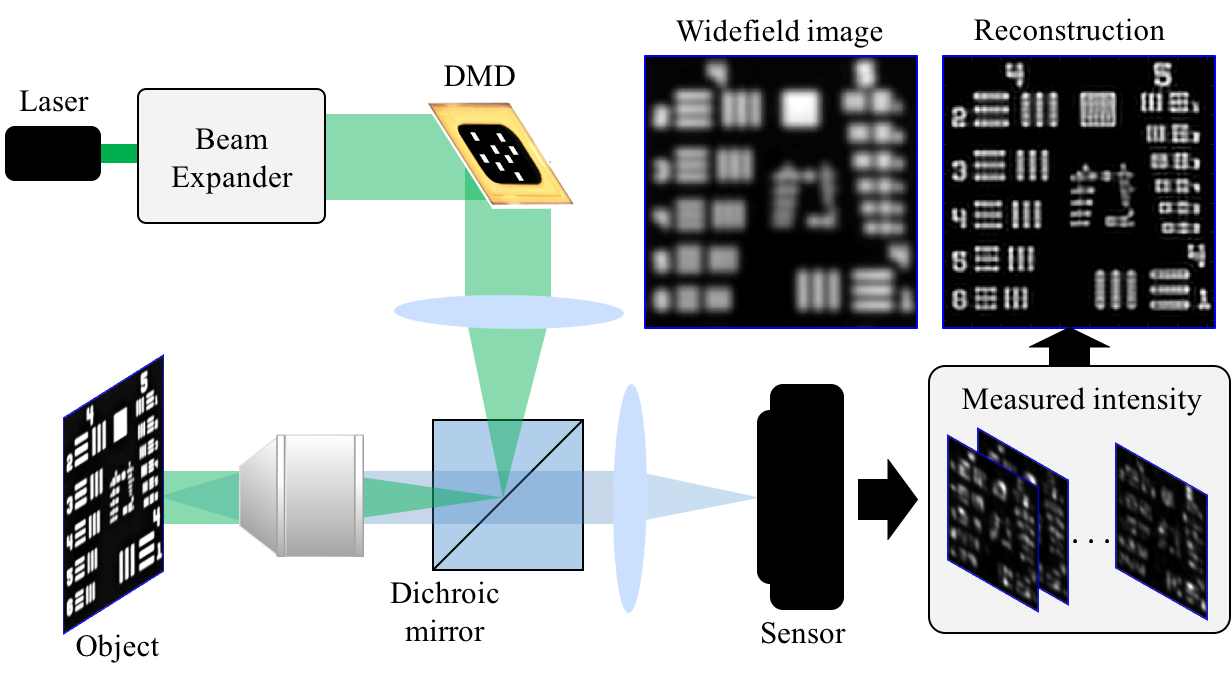}
\caption{Example experimental setup for structured illumination microscopy (SIM) using a deformable mirror device (DMD) to capture low-resolution images of the object modulated by different illumination patterns. Our IPE-SIMS algorithm reconstructs both the super-resoloved image and the unknown arbitrary illumination patterns.}
\label{fig_exp_setup}
\end{figure}

\subsection{Forward model of structured illumination microscopy} \label{sec_setup}

A representative experimental setup is shown in Fig.~\ref{fig_exp_setup}. A DMD spatial light modulator (SLM) is used to project patterns onto the object through an objective lens. The measured intensity for the $\ell$-th captured image is the product of the object's fluorescence distribution $o(\mathbf{r})$ with the illumination pattern $p_\ell(\mathbf{r})$, where $\mathbf{r} = (x,y)$ denotes the lateral position coordinates. This product is then convolved with the system's incoherent detection-side point spread function (PSF), $h_{\mathrm{det}}(\mathbf{r})$: 
\begin{eqnarray}
&&I_\ell(\mathbf{r}) = [o (\mathbf{r}) \cdot p_\ell(\mathbf{r})] \otimes h_{\mathrm{det}}(\mathbf{r}) = \iint o(\mathbf{r}') p_\ell(\mathbf{r}') h_{\mathrm{det}}(\mathbf{r} - \mathbf{r}') \,d^2\mathbf{r}'.
\label{eqn_measurement}
\end{eqnarray}

\subsection{Part 1: Pattern estimation}

\begin{figure}[tbh]
\centering
\includegraphics[width=12cm]{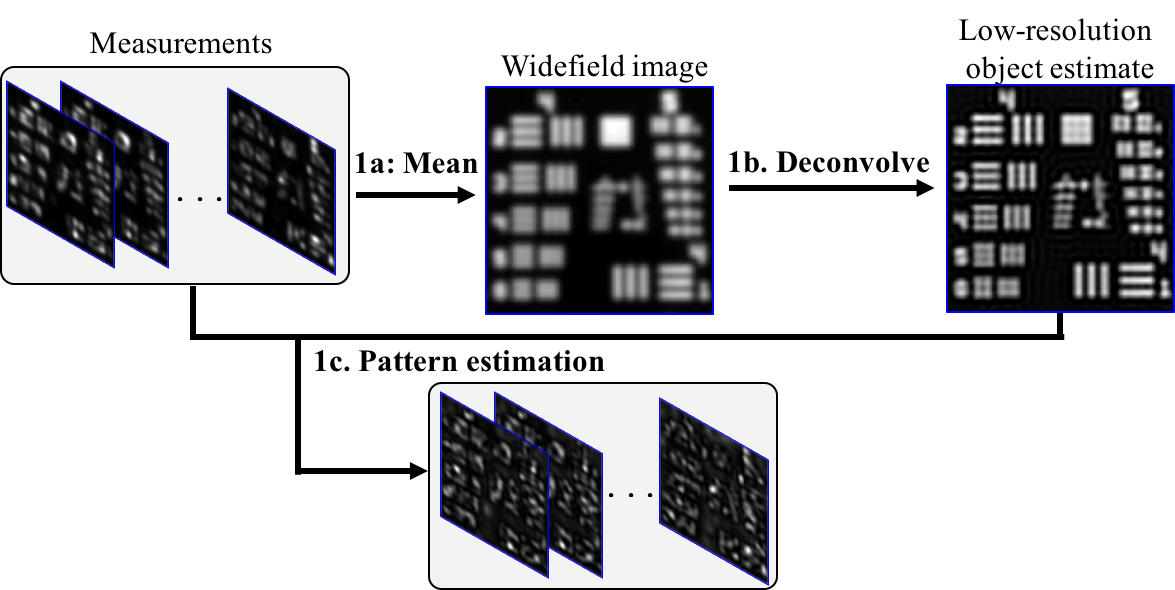}
\caption{The first part of our algorithm, Pattern Estimation (PE), iteratively estimates the illumination patterns from an approximated object given by the deconvolved widefield image.}
\label{fig_IPE}
\end{figure}

The first part of our inverse algorithm is to estimate the illumination patterns. To do so, we start with an low-resolution approximation of the object. Then, we use this object and our measured images to iteratively estimate the patterns (see Fig.~\ref{fig_IPE}).

\subsubsection*{\textbf{Part 1a:} Approximate widefield image}
If we already knew the object $o (\mathbf{r})$, it would be straightforward to estimate the pattern for each measured image by dividing out the object from each of the measurements. However, the object $o(\mathbf{r})$ is unknown. Hence, we start by making a rough estimate of the object. We first take the mean of all the measured images:
\begin{eqnarray}
&&I_{\mathrm{avg}}(\mathbf{r}) = \left< I_\ell (\mathbf{r})\right>_\ell = [o (\mathbf{r}) \cdot \left<p_\ell(\mathbf{r})\right>_\ell] \otimes h_{\mathrm{det}}(\mathbf{r}) \approx p_0 o(\mathbf{r}) \otimes h_{\mathrm{det}}(\mathbf{r}),
\label{eqn_wf_est}
\end{eqnarray}
where $\left<\cdot\right>_\ell$ is the mean operation with respect to $\ell$, and $p_0 = \left< p_\ell(\mathbf{r}) \right>_\ell$ is approximately a constant over the entire field of view. The resulting image will be equivalent to the low-resolution widefield image if the sum of all illumination patterns is approximately uniform.

\subsubsection*{\textbf{Part 1b:} Deconvolve widefield image}

Since the widefield image represents the convolution of the object with its PSF, we can perform a deconvolution operation to estimate the low-resolution object:
\begin{eqnarray}
&&o_{\mathrm{est}}(\mathbf{r}) = \mathcal{F}^{-1} \left\{ \frac{\tilde{I}_{\mathrm{avg}}(\mathbf{u}) \cdot \tilde{h}_{\mathrm{det}}(\mathbf{u})}{|\tilde{h}_{\mathrm{det}}(\mathbf{u})|^2 + \beta}\right\},
\label{eqn_wf_est_dec}
\end{eqnarray}
where $\mathcal{F}$ and $\mathcal{F}^{-1}$ denote the Fourier transform and its inverse, respectively, $\tilde{\cdot}$ denotes the Fourier transform of a certain function, $\mathbf{u} = (u_x, u_y)$ are the lateral spatial frequency coordinates and $\beta$ is a small Tikhonov regularization constant. Note that this object estimate has diffraction-limited resolution and will be used only for estimating the illumination patterns.

\subsubsection*{\textbf{Part 1c:} Pattern estimation}

We then use the low-resolution object estimate $o_{\mathrm{est}}(\mathbf{r})$ to recover each of the illumination patterns. Since each image is simply the product of the illumination and object, we could divide each image by the estimated object to get the pattern. However, we instead solve the problem as an optimization procedure in order to impose the correct Fourier support constraint and avoid reconstruction artifacts. The $\ell$-th pattern estimate is the solution to the following problem
\begin{eqnarray}
\begin{aligned}
& \underset{p_\ell}{\text{minimize}}
& & f(p_\ell) = f_{\mathrm{diff}} (p_\ell) + \mathbb{I}_\mathcal{C}(p_\ell) = \sum_{\mathbf{r}} | I_\ell (\mathbf{r}) - [o_{\mathrm{est}} (\mathbf{r}) \cdot p_\ell(\mathbf{r})] \otimes h_{\mathrm{det}}(\mathbf{r}) |^2 + \mathbb{I}_\mathcal{C}(p_\ell), \\
& \text{where}
& & \mathbb{I}_\mathcal{C}(p_\ell) = \left\{ 
\begin{array}{cl}
0, & p_\ell \in \mathcal{C} \\
+\infty, & p_\ell \notin \mathcal{C}
\end{array}
\right., \;\; \mathcal{C} = \left\{p_\ell(\mathbf{r})\left| \tilde{p}_\ell(\mathbf{u}) = 0, \;\forall \mathbf{u} > \frac{2 NA}{\lambda_{\mathrm{illu}}}\right.\right\},
\end{aligned}
\label{eqn_pattern_est_prox}
\end{eqnarray}
where $\lambda_{\mathrm{illu}}$ is the wavelength of the excitation light.
The first term of the cost function, $f_{\mathrm{diff}}(p_\ell)$, in Eq.~\eqref{eqn_pattern_est_prox} is the least square error (residual) between the measured intensity and the predicted intensity based on our current estimate. The second term enforces a frequency support constraint for the illumination pattern via an indicator function $\mathbb{I}_\mathcal{C}$. This is important to reduce artifacts in the pattern estimation because a normal division between the measured image and estimated object will create errors outside of this frequency support. In our epi-illumination geometry, the constraint is that the frequency content of each illumination pattern be confined within the OTF defined by the objective's NA. 

We implement a proximal gradient descent algorithm~\cite{Parikh2013}, summarized in Subroutine~\ref{algo_pattern}. Proximal gradient descent is designed to solve convex optimization problems like ours that have two cost function terms: one being a differentiable cost function term (e.g. the residual) and the other being a constraint or regularization term (usually nondifferentiable). When the constraint is defined by an indicator function, as in Eq.~\eqref{eqn_pattern_est_prox}, the method is also known as a projected gradient method.

To implement, we first compute the gradient of the differentiable cost function term with respect to $p_\ell(\mathbf{r})$
\begin{eqnarray}
&&\hspace{-0.1 in}g_\ell^{(k)}(\mathbf{r}) = \frac{\partial f_{\mathrm{diff}}(p_\ell^{(k)})}{\partial p_\ell} = -2o_{\mathrm{est}}(\mathbf{r}) \cdot [h_{\mathrm{det}}(\mathbf{r}) \otimes (I_\ell(\mathbf{r}) - [o_{\mathrm{est}} (\mathbf{r}) \cdot p_\ell^{(k)}(\mathbf{r})] \otimes h_{\mathrm{det}}(\mathbf{r}))],
\label{eqn_pattern_grad}
\end{eqnarray}
where $k$ denotes evaluation of the gradient using the pattern at the $k$-th iteration.

We define the projection operation $\Pi_\mathcal{C}$ to force the information outside of the OTF to be zero at each iteration. To reduce high-frequency artifacts, the following soft-edge filter is used
\begin{eqnarray}
&&\Pi_\mathcal{C}(y)  = \mathcal{F}^{-1}\left\{ \frac{\mathcal{F}\{y\} \cdot |\tilde{h}_{\mathrm{illu}}(\mathbf{u})|^2}{|\tilde{h}_{\mathrm{illu}}(\mathbf{u})|^2 + \delta} \right\},
\label{eqn_projection_operator}
\end{eqnarray}
where $h_{\mathrm{illu}}(\mathbf{r})$ is the system's illumination-side PSF, and $\delta$ determines the amount of high-frequency information that is suppressed in the pattern estimation step. We repeat this process of updates and projections until convergence (typically $\sim$50 iterations to estimate each pattern). 

The convergence speed for proximal gradient descent is on the order of $O(1/K)$~\cite{Parikh2013}, indicating that the residual between the current and optimal cost functions is inversely proportional to the number of iterations $K$. To accelerate convergence, one extra step is conducted in Subroutine~\ref{algo_pattern} to include the information of the previous estimate~\cite{Nesterov1983,Beck2009}. The convergence rate for this accelerated proximal gradient method, $O(1/K^2)$~\cite{Beck2009}, is significantly faster than the normal proximal gradient method.

\begin{subroutine}[h]\label{algo_pattern}
 \SetKwInOut{Input}{Input}
 \SetKwInOut{Output}{Output}
 \Input{$I_\ell(\mathbf{r})$, $o_{\mathrm{est}}(\mathbf{r})$}
 initialize $p^{(1)}_\ell(\mathbf{r})$ with all zero image\; $t_1 = 1$\;
 \For{$k = 1:K$}{
  Select step size $\eta^{(k)} > 0$\;
  $\hat{p}_\ell^{(k+1)}(\mathbf{r}) = \Pi_{\mathcal{C}} \left[p_\ell^{(k)}(\mathbf{r}) - \eta^{(k)} g_\ell^{(k)}(\mathbf{r})\right]$, where $\Pi_{\mathcal{C}}$ denotes the projection onto $\mathcal{C}$.
  $t_{k+1} = \frac{1+\sqrt{1+ 4t_k^2}}{2}$ \;
  $p^{(k+1)}_\ell(\mathbf{r}) = \hat{p}_\ell^{(k)}(\mathbf{r}) + \frac{t_k -1}{t_{k+1}} \left[\hat{p}_\ell^{(k+1)}(\mathbf{r}) - \hat{p}_\ell^{(k)}(\mathbf{r})\right]$\;
 }
 \Output{$p_\ell(\mathbf{r})$}
 \caption{Pattern Estimation}
\end{subroutine} 

\subsection{Part 2: SIM with a statistical prior} \label{sec_SIMS}

\begin{figure}[h]
\centering
\includegraphics[width=13cm]{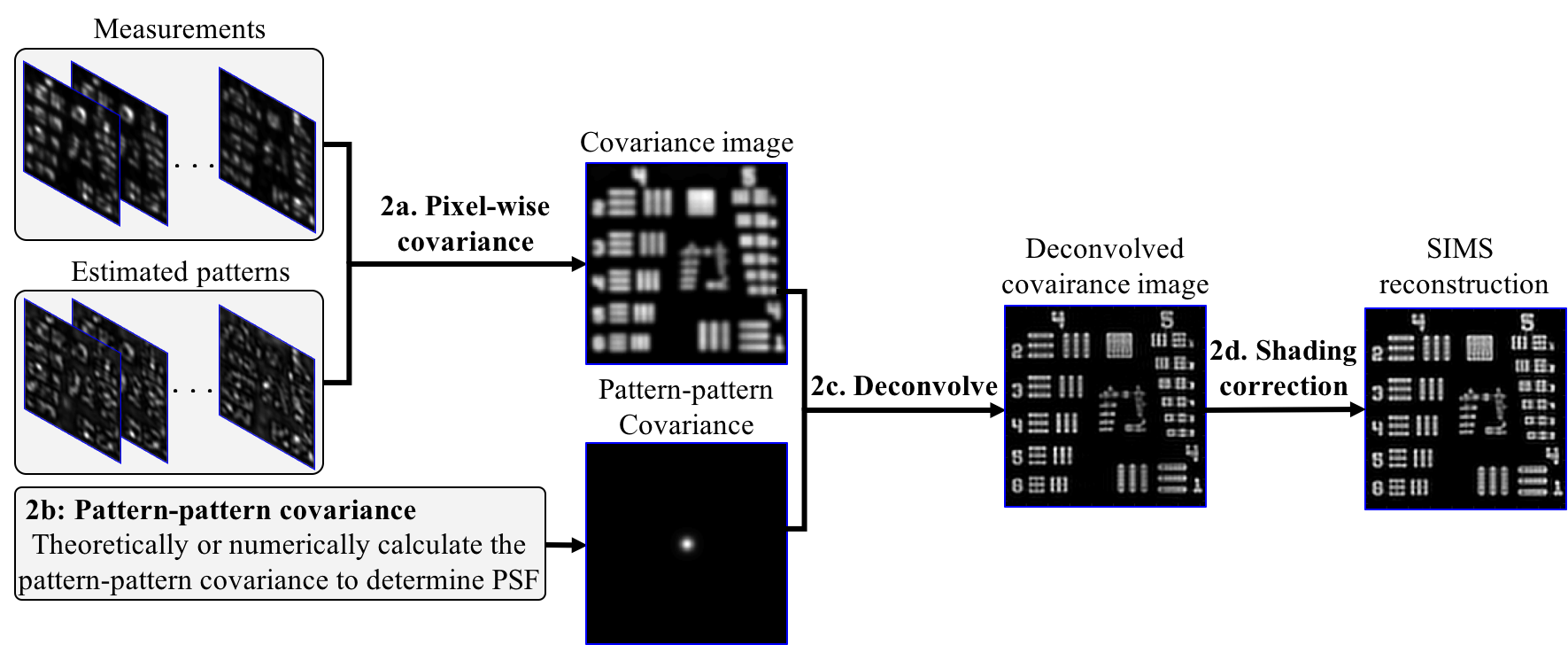}
\caption{The second part of our algorithm, termed structured illumination microscopy with a statistical prior (SIMS), estimates the high-resolution object from the measured images and the estimated illumination patterns obtained in Part 1.}
\label{fig_SIMS}
\end{figure}

Once we have recovered the illumination patterns, the second part of the algorithm is to reconstruct a high-resolution image from the measured dataset $I_\ell(\mathbf{r})$ and the estimated patterns $p_\ell(\mathbf{r})$. We call this part of the algorithm Structured Illumination Microscopy with a Statistical prior (SIMS), summarized in Fig.~\ref{fig_SIMS}. There are four steps, which are explained below. We will also describe how the statistical prior is used and why this procedure gives better resolution.  

\subsubsection*{\textbf{Part 2a:} Calculate the pattern-intensity covariance}

Consider the case where the pattern $p(\mathbf{r})$ is a random variable at position $\mathbf{r}$ and the measured intensity $I(\mathbf{r})$ is also a random variable at position $\mathbf{r}$. The $\ell$-th image is thus the $\ell$-th sample function for these random variables (one event out of the sample space). Covariance is a measure of how much two random variables change together. Since the intensity $I(\mathbf{r})$ is the blurred version of the product between random patterns $p(\mathbf{r})$ and deterministic object $o(\mathbf{r})$ (Eq.~\eqref{eqn_measurement}), the covariance between the pattern and the intensity should give high similarity wherever the object $o(\mathbf{r})$ has signal and thus allow us to find the object underneath the random-pattern illumination~\cite{Tanaami1996, Walker2001, Jiang2004, Garcia2005,Heintzmann2006}. We calculate this covariance image $I_{\mathrm{cov}}(\mathbf{r})$ as 
\begin{eqnarray}
&&I_{\mathrm{cov}}(\mathbf{r}) = \left< \Delta p_\ell(\mathbf{r}) \Delta I_\ell(\mathbf{r}) \right>_\ell = \iint o(\mathbf{r}') \left< \Delta p_\ell (\mathbf{r}) \Delta p_\ell(\mathbf{r}')\right>_\ell h_{\mathrm{det}}(\mathbf{r} - \mathbf{r}') d^2 \mathbf{r}',
\label{eqn_IPE_SIM}
\end{eqnarray}
where $\Delta I_\ell(\mathbf{r}) = I_\ell(\mathbf{r}) - \left< I_\ell(\mathbf{r})\right>_\ell$, and $\Delta p_\ell(\mathbf{r}) = p_\ell(\mathbf{r}) - \left< p_\ell(\mathbf{r})\right>_\ell$.

Regardless of which illumination pattern is imposed, the covariance image always gives an estimate of the object. However, the resolution of the reconstructed object may be different for different pattern statistics. We can quantify this by taking a closer look at the expression on the right-hand side of Eq.~\eqref{eqn_IPE_SIM}. The multiplication of detection PSF and covariance between $p(\mathbf{r})$ and $p(\mathbf{r}')$ acts as the PSF of the covariance image, which thus determines resolution. If the patterns are perfectly spatially correlated, the pattern-pattern covariance is a constant, and the pattern-intensity covariance image is a normal widefield image with PSF of $h(\mathbf{r})$. If the patterns are perfectly spatially uncorrelated, the pattern-pattern covariance is $\left<|\Delta p_\ell(\mathbf{r})|^2 \right>_\ell \delta(\mathbf{r} - \mathbf{r}')$, which, for a constant variance, results in the PSF being a delta function and the object being reconstructed with perfect resolution. In practice, this is not achievable, since the illumination is bandlimited and thus cannot be perfectly uncorrelated. In the general case, to find the resolution (PSF) of the covariance image, we need to calculate the spatial covariance of the patterns, which is the subject of Part 2b, below.

\subsubsection*{\textbf{Part 2b:} Calculate pattern-pattern covariance}

To calculate the spatial covariance of the projected patterns, we first consider the pattern formation model. In our experiments, for example, we use a DMD to create random patterns at the sample plane. Assuming that the projected DMD pattern is sparse enough to avoid interference cross-terms, we can express our pattern under the incoherent model as
\begin{eqnarray}
&&p_\ell(\mathbf{r}) = \iint t_\ell(\mathbf{r}') h_{\mathrm{illu}}(\mathbf{r} - \mathbf{r}') d^2 \mathbf{r}',
\label{eqn_DMD_pattern}
\end{eqnarray}
where $t_\ell(\mathbf{r})$ is the $\ell$-th pattern on the DMD. With this model, the pattern-pattern covariance is
\begin{eqnarray}
&&\left< \Delta p_\ell(\mathbf{r}) \Delta p_\ell(\mathbf{r}')\right>_\ell = \iint \iint \left< \Delta t_\ell(\mathbf{r}_1) \Delta t_\ell(\mathbf{r}_2) \right>_\ell h_{\mathrm{illu}}(\mathbf{r} - \mathbf{r}_1) h_{\mathrm{illu}}(\mathbf{r}' - \mathbf{r}_2) d^2 \mathbf{r}_1 d^2 \mathbf{r}_2 \nonumber \\
&&\hspace{1 in} = \iint \iint \gamma_t\left< \Delta t_\ell^2(\mathbf{r}_1)\right>_\ell \delta(\mathbf{r}_1 - \mathbf{r}_2) h_{\mathrm{illu}}(\mathbf{r} - \mathbf{r}_1) h_{\mathrm{illu}}(\mathbf{r}' - \mathbf{r}_2) d^2 \mathbf{r}_1 d^2 \mathbf{r}_2 \nonumber \\
&&\hspace{1 in} \approx \alpha_t \iint h_{\mathrm{illu}}(\mathbf{r} - \mathbf{r}_1) h_{\mathrm{illu}}(\mathbf{r}' -\mathbf{r}_1) d^2\mathbf{r}_1= \alpha_t (h_{\mathrm{illu}} \star h_{\mathrm{illu}}) (\mathbf{r} - \mathbf{r}'),
\label{eqn_DMD_correlation}
\end{eqnarray}
where we have used an assumption that the DMD pattern values at position $\mathbf{r}_1$ and $\mathbf{r}_2$ are perfectly uncorrelated:
\begin{eqnarray}
&&\left< \Delta t_\ell(\mathbf{r}_1)\Delta t_\ell(\mathbf{r}_2)\right>_\ell = \gamma_t \left< \Delta t_\ell^2(\mathbf{r}_1)\right>_\ell \delta(\mathbf{r}_1 - \mathbf{r}_2) \approx \alpha_t \delta(\mathbf{r}_1 - \mathbf{r}_2),
\label{eqn_correlation_assumption}
\end{eqnarray}  
with $\gamma_t$ being a constant that maintains unit consistency. This assumption is valid because the effective DMD pixel size is small compared to the FWHM of the optical system and we can control $\Delta t_\ell(\mathbf{r})$ to create an uncorrelated pattern. In the experiment, each position of $t_\ell(\mathbf{r})$ is an independent and identically distributed random variable. When the number of patterns is large enough, the variance $\left< \Delta t_\ell^2(\mathbf{r}_1)\right>_\ell$ approaches the same constant for all the positions. We can then combine $\gamma_t$ and the variance into a single constant $\alpha_t$. 

Ideally, we can assume $h_{\mathrm{illu}}(\mathbf{r}) \approx h_{\mathrm{det}}(\mathbf{r})$ when $\lambda_{\mathrm{illu}} \approx \lambda_{\mathrm{det}}$, where $\lambda_{\mathrm{det}}$ is the wavelength of the fluorescent emission detection light, and theoretically calculate the pattern-pattern covariance. We can also estimate $h_{\mathrm{illu}} \star h_{\mathrm{illu}}(\mathbf{r})$ by numerically evaluating Eq.~\eqref{eqn_DMD_correlation} using our estimated patterns, which accounts for possible aberrations in the illumination optics.

\subsubsection*{\textbf{Part 2c:} PSF deconvolution of the covariance image}

The pattern-pattern covariance derived in Part 2b is related to the PSF of the pattern-intensity covariance calculated in Part 2a. Hence, we can plug the pattern-pattern covariance into Eq.~\eqref{eqn_IPE_SIM} and write the covariance image as
\begin{eqnarray}
&&I_{\mathrm{cov}}(\mathbf{r}) = \left< \Delta p_\ell(\mathbf{r}) \Delta I_\ell(\mathbf{r}) \right>_\ell =  \iint \alpha_t o(\mathbf{r}') [(h_{\mathrm{illu}}\star h_{\mathrm{illu}})\cdot h_{\mathrm{det}}](\mathbf{r} - \mathbf{r}') d^2 \mathbf{r}'.
\label{eqn_IPE_SIM_DMD}
\end{eqnarray}

Importantly, the effective PSF for this correlation image is now $[(h_{\mathrm{illu}} \star h_{\mathrm{illu}}) \cdot h_{\mathrm{det}}] (\mathbf{r})$, and the corresponding effective OTF is $[|\tilde{h}_{\mathrm{illu}}|^2 \otimes \tilde{h}_{\mathrm{det}}](\mathbf{u})$. Since both $|\tilde{h}_{\mathrm{illu}}|^2$ and $\tilde{h}_{\mathrm{det}}$ have approximately the same Fourier support as the widefield OTF, the convolution between them covers around $2\times$ the support of the widefield OTF, as in conventional SIM. Given the effective PSF, we implement a standard deconvolution to improve contrast at high spatial frequencies: 
\begin{eqnarray}
&&I_{\mathrm{cov,dec}} (\mathbf{r}) = \mathcal{F}^{-1}\left\{ \frac{\tilde{I}_{\mathrm{cov}}(\mathbf{u}) \cdot H(\mathbf{u})}{|H(\mathbf{u})|^2 + \xi} \right\},
\label{eqn_IPE_SIM_DMD_dec}
\end{eqnarray}
where $H(\mathbf{u}) = [|\tilde{h}_{\mathrm{illu}}|^2 \otimes \tilde{h}_{\mathrm{det}}](\mathbf{u})$ and $\xi$ is a small regularization parameter. 

\subsubsection*{\textbf{Part 2d:} Shading correction operation}

When the number of images is not large enough to give uniform variance of the patterns at each pixel ($\left< \Delta t_\ell^2(\mathbf{r}')\right>_\ell$ from Eq.~\eqref{eqn_DMD_correlation}), low-frequency shading artifacts will occur. Even if we assume the mean of the pattern to be flat in Eq.~\eqref{eqn_wf_est}, the variance can still be non-uniform. These can be seen in the deconvolved covariance image in Fig.~\ref{fig_SIMS}. To resolve this, we can estimate and correct for the variance across the image using our previously estimated projected patterns. Since the projected pattern $p_\ell(\mathbf{r})$ is the blurred version of the pattern on the DMD, by ignoring the high-frequency component of the DMD pattern, we can approximate the variance of the DMD pattern by 
\begin{eqnarray}
&&\alpha_t(\mathbf{r}) = \gamma_t \left< \Delta t_\ell^2(\mathbf{r})\right>_\ell \approx \gamma_t \left< \Delta p_\ell^2(\mathbf{r})\right>_\ell.
\label{eqn_variance_approx}
\end{eqnarray}
We divide out the spatially-varying variance $\alpha_t$ in Eq.~\eqref{eqn_IPE_SIM_DMD} from the deconvolved SIMS image,
\begin{eqnarray}
&&I_{\mathrm{SIMS}}(\mathbf{r}) = \frac{I_{\mathrm{cov,dec}}(\mathbf{r}) \cdot \alpha_t (\mathbf{r})}{ \alpha_t^2 (\mathbf{r}) + \epsilon},
\label{eqn_shading_correction}
\end{eqnarray}
where $\epsilon$ is a regularizer and $I_{\mathrm{SIMS}}(\mathbf{r})$ is the output from our SIMS reconstruction (Part 2c). This result of this step is our final reconstruction of the high-resolution object function. 

\subsection{Parameter Tuning and Algorithm Runtime}

Our SIMS algorithm involves 4 regularizers: $\beta$, $\delta$, $\xi$, and $\epsilon$, described in Eq.~\eqref{eqn_wf_est_dec}, Eq.~\eqref{eqn_projection_operator}, Eq.~\eqref{eqn_IPE_SIM_DMD_dec}, and Eq.~\eqref{eqn_shading_correction}, respectively. Each is decoupled from the others and acts similarly to a typical Tikhonov regularizer, so tuning may be done independently. Generally, we want the regularizers to be as small as possible, while still avoiding noise amplification. 

The procedure to tune the regularization parameters heuristically is summarized as follows. First, we check if the widefield images are well-deconvolved by finding the smallest $\beta$ to give the image with best resolution but without obvious noise amplification, then we move on to check the deconvolved covariance image by tuning the SIMS regularizer $\xi$ and the smooth-edge filter regularizer $\delta$ using the same principle, and finally we check the final reconstruction by using the smallest shading correction regularizer $\epsilon$ with enough shading correction but without evident noise amplification. Additionally, the negative values in all of the deconvolved images are set to zero since the fluorescent density is always positive.

The algorithm is implemented in MATLAB and run on an Intel i7 2.8 GHz CPU computer with 16 G DDR3 RAM under OS X operating system. To reconstruct an image with size of $200\times 200$ pixels and 400 measurements, this computer takes about 200 seconds. The bottleneck of the algorithm is on the pattern estimation step. The estimation of each pattern takes around 0.5 second.

\section{Results} \label{Sec: Def_of_res}

\subsection{Definition of resolution}

Before introducing and comparing any SIM algorithms, we want to first define the resolution criterion considered in this paper. Resolution of a microscopic image is usually defined by measuring the minimal resolvable distance between two points. Consider a widefield image with detected wavelength $\lambda$ and numerical aperture $NA$; the Abbe resolution criterion is then $0.5\lambda/NA$, the full width at half maximum (FWHM) of the widefield PSF. As two points get closer to each other, the contrast between them decreases. Under the separation set by Abbe's limit, two infinitely small points observed under widefield microscope will give an overlapped two-point image with a dip at the center with the contrast equal to 0.01. Hence, the Abbe resolution criterion can be thought of as setting the minimum acceptable contrast between two points at 0.01. We can therefore define the resolution of a microscope or a reconstruction algorithm by measuring the smallest resolvable fine features that have contrast between them of at least 0.01. 

\subsection{Comparison of algorithms}
\begin{figure}[bth]
\centering
\includegraphics[width=13cm]{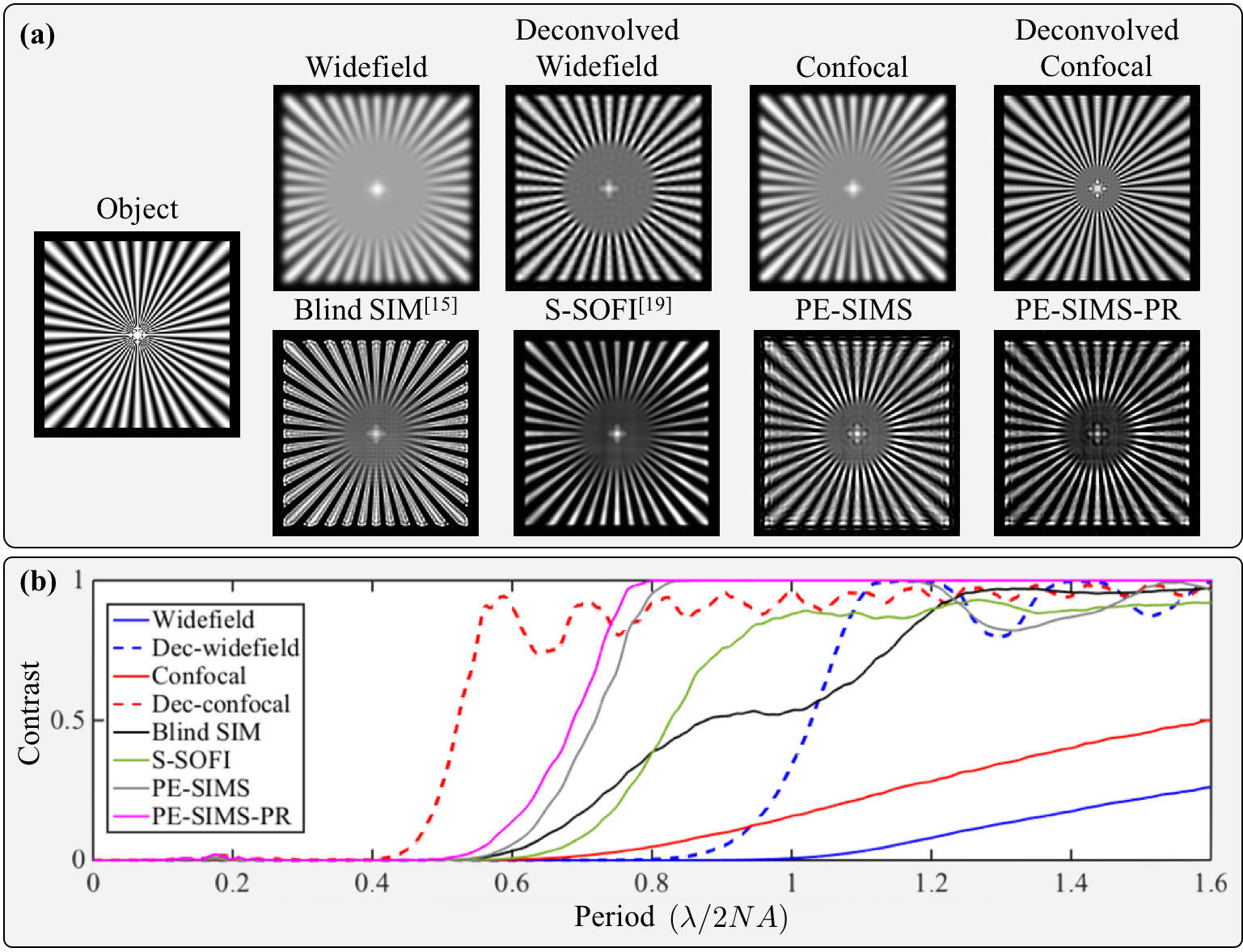}
\caption{(a) Simulated reconstructions of a Siemens star target under a widefield microscope, deconvolved widefield, confocal microscope, deconvolved confocal, blind SIM~\cite{Mudry2012}, S-SOFI~\cite{Kim2015}, our PE-SIMS and PE-SIMS-PR algorithms. (b) The effective modulation transfer function (MTF) of each method, given by the contrast of the reconstructed Siemens star image at different radii. }
\label{fig_res_def}
\end{figure}

Given this definition of resolution, we quantify the resolution for various algorithms in Fig.~\ref{fig_res_def}. The Siemens star test target ($o(r,\theta) = 1+\cos 40\theta$ in polar coordinates) has varying spatial frequencies along the radius. The resolution of different imaging methods is quantified by reading the minimal resolved period when the contrast reaches 0.01. The effective modulation transfer function (MTF) of each method is shown in Fig.~\ref{fig_res_def}b, measured as the contrast of the reconstructed Siemens star image at different radii.

Our simulations use a SIM dataset with random patterns, so that we may compare against the previously proposed reconstruction algorithms of blind SIM~\cite{Mudry2012} and S-SOFI~\cite{Kim2015}. We create $N_{\mathrm{img}} = 400$ speckle-illuminated images from shifted random patterns on the DMD, with shifts of $0.6$ FWHM of the PSF across $20 \times 20$ steps in the $x$ and $y$ directions, respectively. In each pattern, only $10\%$ of the DMD pixels are turned on. This noise-free situation allows us to compare the ideal achieved resolution for the different algorithms. 

Figure~\ref{fig_res_def}a shows the widefield, deconvolved widefield, confocal, and deconvolved confocal images of the Siemens star, as compared to blind SIM~\cite{Mudry2012}, S-SOFI~\cite{Kim2015} and our algorithm. At the bottom, we show the measured effective MTF for each algorithm. In terms of visual effect, S-SOFI~\cite{Kim2015} gives the least artifacts. 

\begin{table}[bth]
\caption{Achieved resolution for different algorithms}
\centering
\begin{tabular}{|c|c|c|c|c|}
\hline
  & Widefield & \makecell{Widefield \\ deconvolved} & Confocal & \makecell{Confocal \\ deconvolved} \\ \hline
\makecell{Resolution \\ $[\lambda/2NA]$} & 1.035 & 0.844 & 0.681 & 0.428 \\ \hline
Enhancement & 1 $\times$ & 1.23 $\times$ & 1.52 $\times$ & 2.42 $\times$ \\ \hline
\hline 
   & Blind SIM & S-SOFI & PE-SIMS & PE-SIMS-PR \\ \hline
\makecell{Resolution \\  $[\lambda/2NA]$} & 0.563  & 0.619 & 0.551 & 0.517 \\ \hline
Enhancement & 1.84 $\times$ & 1.67 $\times$ & 1.88 $\times$ & 2.00 $\times$\\ \hline
\end{tabular}
\label{tab_achieve_res}
\end{table}

To compare resolution, we use our definition of the minimal resolved separation when the contrast drops to 0.01 and summarize the results in Table~\ref{tab_achieve_res}. The enhancement metric gives the ratio resolution improvement over widefield imaging. S-SOFI resolves features down to 1.67 $\times$ smaller than the widefield microscope, which is close to the claimed 1.6$\times$ in~\cite{Kim2015}, and Blind SIM achieves $1.84 \times$ improvement but lower contrast for high-frequencies, which is consistent with~\cite{Mudry2012}. Our PE-SIMS and PE-SIMS-PR (PE-SIMS with pixel reassignment algorithm ~\cite{Cox1982, Sheppard1988, Muller2010, Roth2013, Sheppard2013} described in Appendix B) algorithms give better resolution compared to other methods. We resolve features down to $1.84\times$ and $2\times$, respectively, close to the limit set by the deconvolved confocal image. Hence, our method performs the best of the blind algorithms.

Ideally, if we know all the patterns and our spatial modulation covers the full Fourier bandwidth of the objective, we could reconstruct out to $4NA/\lambda$ in Fourier space, achieving enhancement of $2.42\times$, as in the case of deconvovled confocal image or periodic SIM with known patterns. The blind algorithms, however, deal with an ill-posed problem (measure $N_{\mathrm{img}}$ images and solve $N_{\mathrm{img}}+1$ images) that can only become well-posed through appropriate constraints. If the prior for these algorithms are not accurate enough, they may solve a different problem even if the problem becomes well-posed. This is why algorithms with different prior assumptions give different resolution performance for the same dataset, as we saw in Table~\ref{tab_achieve_res}.

\section{Experimental Results}

Our experimental setup is shown in Fig.~\ref{fig_exp_setup}. A laser beam (Thorlabs, CPS532, 4.5 mW) is expanded to impinge onto a reflective DMD spatial light modulator (DLP\textregistered Discovery 4100, .7" XGA, 1024$\times$768 pixels, pixel size 13.6 $\mu$m). The DMD generates a total of $N_{\mathrm{img}}$ random patterns ($30\%$ of DMD pixels turned on). These random illumination patterns are projected onto the object (with demagnification of 60$\times$) through a 4\textit{f} system composed of a 200 mm convex lens and a $60 \times$ objective lens with NA$ = 0.8$ (Nikon CFI). The resulting fluorescent light is then collected with another 4\textit{f} system formed by the same $60 \times$ objective and a 400 mm convex lens (magnification $120\times$). A dichroic mirror blocks the reflected illumination light (as in a typical epi-illumination setup). The images are taken with an sCMOS camera (PCO.edge 5.5, 2560$\times$2160 pixels, pixel size 6.5 $\mu$m). Patterns are shifted on a $20\times 20$ grid in the $x$ and $y$ directions with a step size of $0.6$ FWHM of the PSF, while collecting images at each step. Our test object is carboxylate-modified red fluorescent beads (Excitation wavelength: 580 nm/Emission wavelength: 605 nm) having mean diameter of 210 nm (F8810, Life Technologies). 

\begin{figure}[tbh]
\centering
\includegraphics[width=14cm]{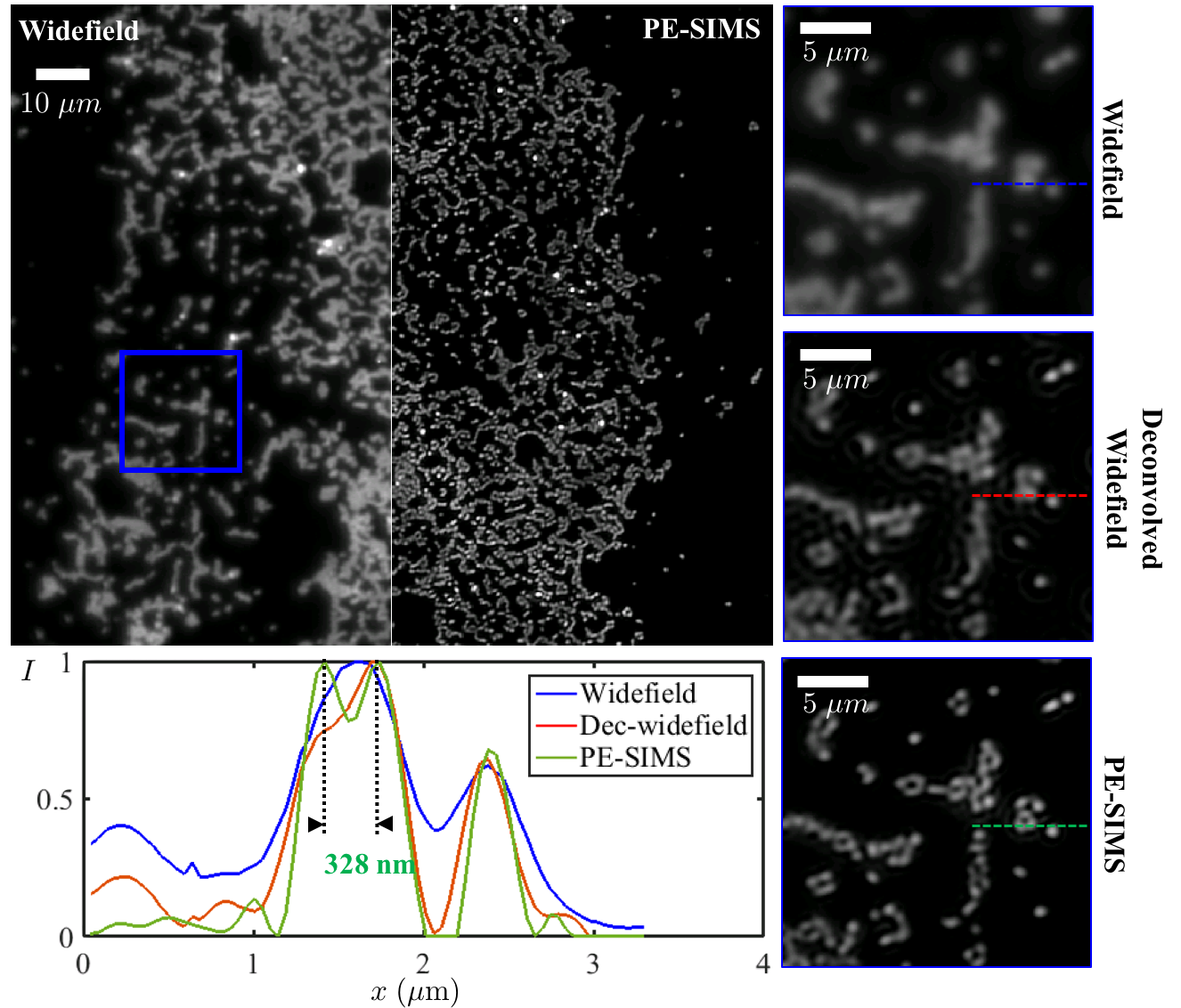}
\caption{Reconstructions of red fluorescent beads (Ex:580 nm/Em:605 nm) from the experiment using random pattern illumination with $20\times 20$ scanning step.}
\label{fig_exp_result}
\end{figure}

Reconstruction results are shown in Fig.~\ref{fig_exp_result}, demonstrating improved resolution using our PE-SIMS algorithm, as compared to standard widefield or deconvolved widefield images. 

To quantitatively analyze the experimental results, we measure the resolved feature size of the reconstructed image and compare it to our theory. As shown in the cutline in Fig.~\ref{fig_exp_result}, two fluorescent beads separated by 328 nm can clearly be resolved using our method, which are otherwise unresolvable in either widefield or deconvolved widefield images. The contrast of this two-Gaussian shape shows these two Gaussian are separated by 1.16$\times$ FWHM, so the FWHM of the reconstructed beads is around 283 nm. Assuming the bead can be modeled as a Gaussian function with FWHM of 140 nm (210 nm in diameter for the beads), we can then deconvolve the bead shape out of the reconstruction and get the FWHM of the PSF for this case equal to 240 nm, which is below the diffraction limit $\lambda/2NA = 371$ nm.  

\begin{figure}[h]
\centering
\includegraphics[width=13cm]{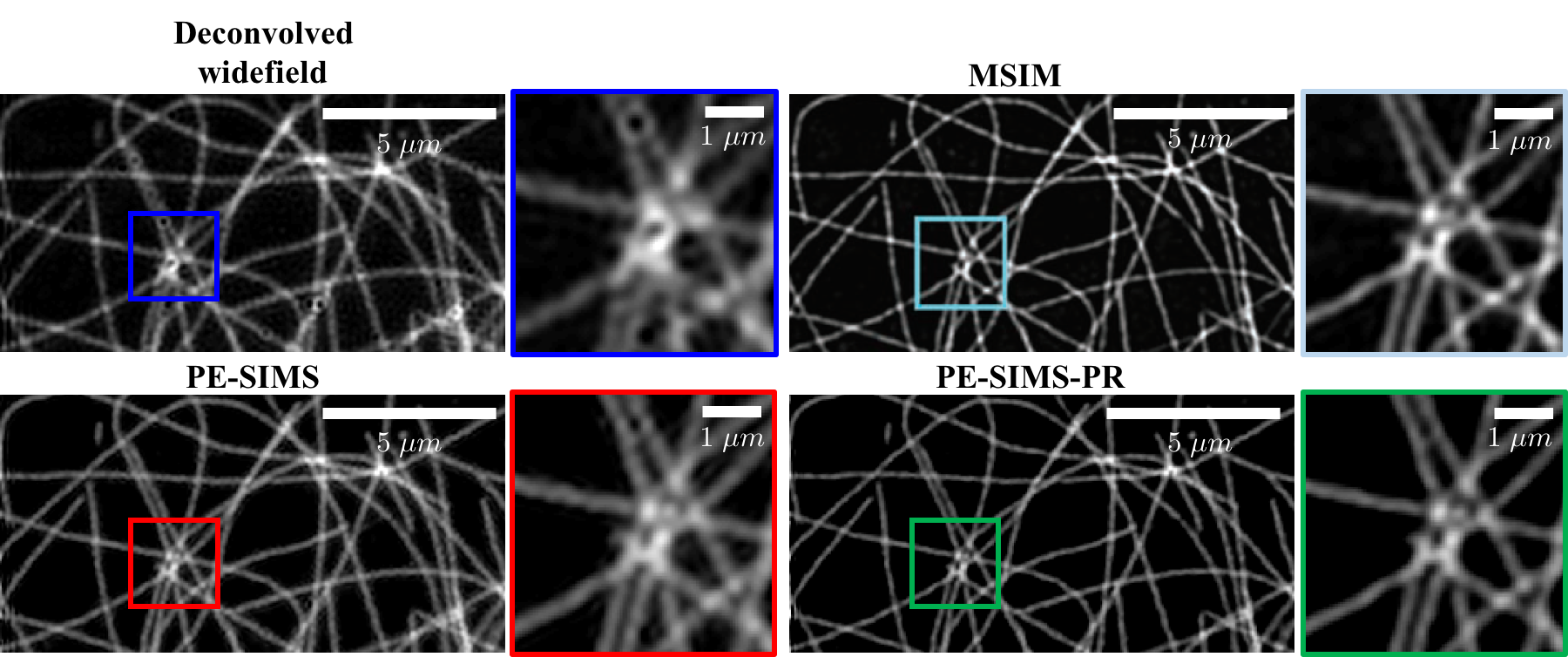}
\caption{Comparison of our algorithm on dataset from Multispot SIM (MSIM) which uses with$N_{\mathrm{img}} = 224$ scanned multi-spot patterns from~\cite{York2012}. We show the deconvolved widefield image and the reconstructions using MSIM with known patterns, as well as our blind PE-SIMS algorithm with and without pixel reassignment.}
\label{fig_SIMS_PR_Nat_data}
\end{figure}

Our algorithm can be used on other types of SIM datasets, as long as the pattern-pattern covariance gives a point-like function at the center. As an example, we tested our algorithm on a dataset from a previous method, Multispot SIM (MSIM)~\cite{York2012}. In MSIM, the patterns are a shifting grid of diffraction-limited spots. Since the previous MSIM implementation assumes known patterns, a calibration step captured an extra dataset with a uniform fluorescence sample in order to measure the patterns directly. Our algorithm ignores this calibration data, yet accurately reconstructs both the object and patterns (see Fig.~\ref{fig_SIMS_PR_Nat_data}). The MSIM result using the calibration data is shown for comparison. The sample is microtubules stained with Alexa Fluor 488 in a fixed cell observed under a TIRF $60\times$ objective with $NA = 1.45$. Our PE-SIMS-PR reconstruction gives a similar result to the known-pattern MSIM reconstruction.

\section{Conclusion}

We have proposed a robust algorithm that can give 2$\times$ resolution improvement compared to widefield fluoresence imaging using a SIM dataset without knowing the imposed patterns. Our algorithm first estimates each illumination pattern from a low-resolution approximate object and measured intensities by solving a constrained convex optimization problem. We then synthesize a high-resolution image by calculating the covariance between the estimated patterns and the measured intensity images, followed by a deconvolution and shading correction to get to the final reconstruction. We quantified the limits on resolution of our algorithm by the reconstructed contrast of a simulated Siemens star target. In simulations, we showed that our algorithm gives better resolution compared to previously proposed blind algorithms~\cite{Mudry2012, Kim2015}. Experimentally, we demonstrated this improvement experimentally on both random speckle pattern illumination and multi-spot scanned illumination.

\section*{Acknowledgment}

The authors thank Prof. Michael Lustig, Jonathan Tamir, Michael Chen and Hsiou-Yuan Liu for helpful discussions. This research is funded by the Gordon and Betty Moore Foundation's Data-Driven Discovery Initiative through Grant GBMF4562 to Laura Waller (UC Berkeley).

\section*{Appendix A: Reducing the number of images by multi-spot scanning}

In this paper, we used 400 random speckle illumination patterns to reconstruct the image, far more than the 9-image requirement of conventional SIM~\cite{Gustafsson2000}. This large number of images was required for high-quality reconstructions because the average and variance of the illumination patterns must be sufficiently flat in order to avoid shading variations. Recall that we want $\alpha_t(\mathbf{r}) \approx \gamma_t \left<\Delta p_\ell^2 (\mathbf{r})\right>_\ell$ in Eq.~\eqref{eqn_variance_approx} to be close to a constant, which suggests that the variance of the random patterns is constant. When the number of images $N_{\mathrm{img}}$ goes down, this statistical assumption is not true any more. We use a shading correction algorithm (Sec.~\ref{sec_SIMS}) to fix this problem by estimating the nonuniform variance, but it is still only an estimate. Hence, when the degree of variance nonuniformity increases (as the number of images decresases), the shading correction algorithm incurs errors. 

\begin{figure}[tbh]
\centering
\includegraphics[width=12cm]{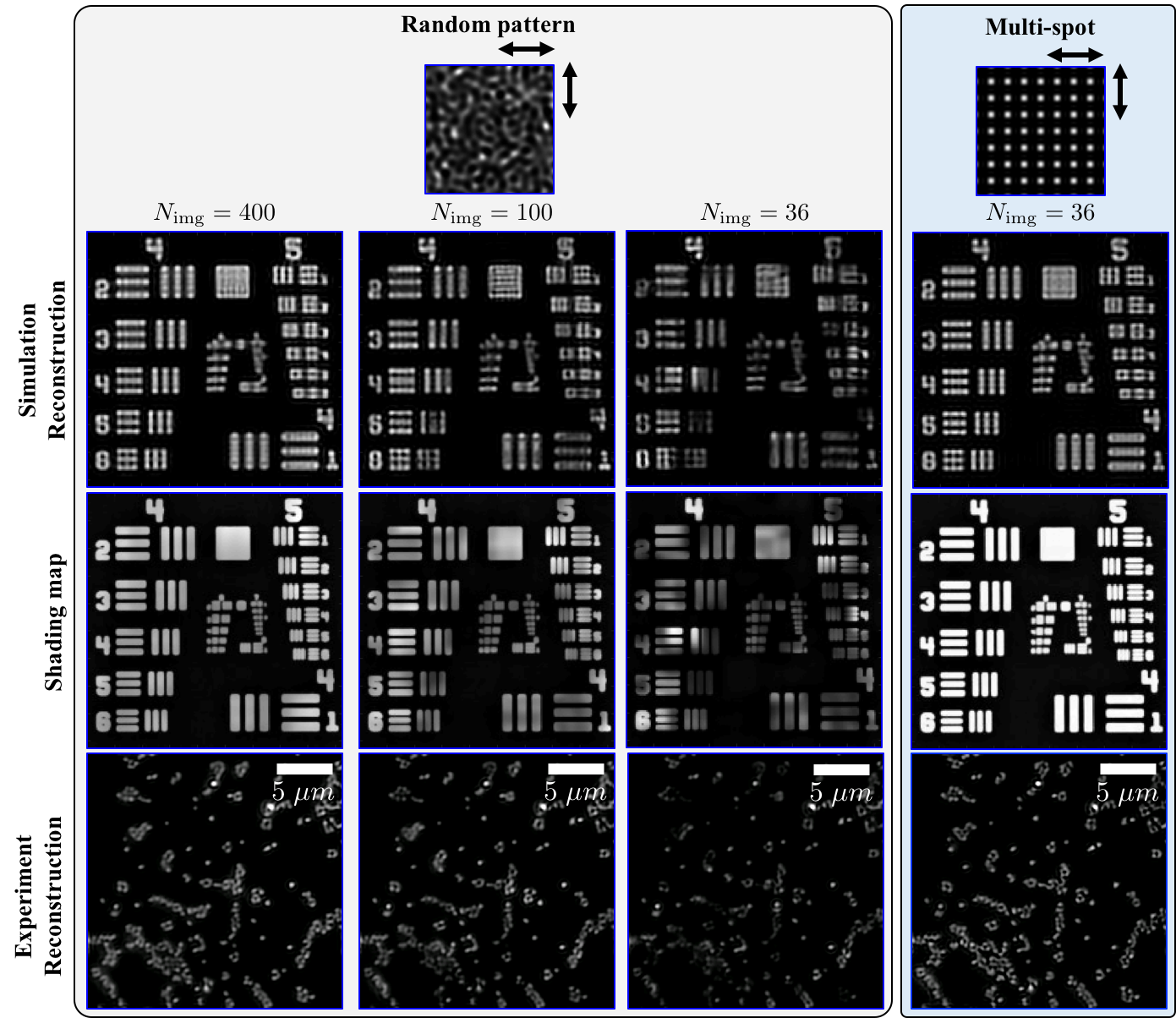}
\caption{Results with simulated and experimental (fluorescent beads) datasets comparing random speckle and multi-spot illumination patterns. (middle row) Shading maps overlaid on the object. Decreasing the number of random patterns results in shading artifacts in the reconstruction. The random patterns are scanned in $20\times 20$, $10\times 10$, and $6\times 6$ steps with the same step size of 0.6 FWHM of the PSF, while the multi-spot pattern is scanned with $6\times 6$ steps. }
\label{fig_scanning_method}
\end{figure}

Figure~\ref{fig_scanning_method} shows simulations demonstrating the effect of reducing the number of images. We use the same random pattern as in Sec.~\ref{sec_SIMS} and shift by step sizes of 0.6 FWHM of the PSF. As we decrease the number of images from 400 to 36, the reconstruction becomes worse, due to shading errors. The shading map, $\alpha_t(\mathbf{r}) o(\mathbf{r})$, is shown in the bottom row of Fig.~\ref{fig_scanning_method}. We can see the artifacts happen at the region where the $\alpha_t(\mathbf{r})$ is dim and changing. Without knowing the patterns a priori it is not possible to fully correct these shading effects.

Since we know that the artifacts that appear with too few images are due to a non-uniform $\alpha_t(\mathbf{r})$, we can attempt to design patterns that will be uniform with a minimal number of images. We would like $\left<\Delta p_\ell^2(\mathbf{r})\right>_\ell$ to give a uniform map. Consider the contribution from a single pattern; $\Delta p^2_\ell(\mathbf{r})$ is similar to the original pattern but with sharper bright spots. The ensemble average over $\ell$ sums up all these bright spots after shifting the pattern around. For a shifted random pattern, we must capture many images in order for the summation of the bright spots to give a uniform map. One efficient way to get a sum of bright spots to become a uniform map is to use a periodic multi-spot pattern (see Fig.~\ref{fig_scanning_method})~\cite{York2012,York2013,Chakrova2015}. The period of this multi-spot pattern is designed to be 6 shifting step sizes. Thus, we can use $6\times 6$ scanning steps to give a uniform shading map $\alpha_t(\mathbf{r})$. The reconstruction is also shown in Fig.~\ref{fig_scanning_method} to be almost as good as the one illuminated with $400$ shifted random patterns. 

Experimentally, we see similar trends in image reconstruction quality for different illumination strategies (see the bottom row of Fig.~\ref{fig_scanning_method}). Results from random pattern illumination of fluorescent beads with $N_{\mathrm{img}} = 400$ and multi-spot illumination with $N_{\mathrm{img}} = 36$ give very similar results, and shading artifacts become prominent as the number of patterns is reduced. Note that we use the same algorithm for both the random and multi-spot illuminated datasets because the PSFs of the pattern-intensity covariance images $I_{\mathrm{cov}}(\mathbf{r})$ for both cases are the same.

To show that the PSF for the pattern-intensity covariance image with random and multi-spot illumination are the same, we must derive the pattern-pattern covariance $\left< \Delta p_\ell(\mathbf{r}) \Delta p_\ell(\mathbf{r}') \right>_\ell$ as we did in Part 2b in Sec.~\ref{sec_SIMS}. To calculate the pattern-pattern covariance, we need to calculate the covariance of the patterns on the DMD $\left<\Delta t_\ell(\mathbf{r}) \Delta t_\ell(\mathbf{r}')\right>_\ell$ and plug it into Eq.~\eqref{eqn_DMD_correlation} to get pattern-pattern covariance $\left<\Delta p_\ell(\mathbf{r}) \Delta p_\ell(\mathbf{r}')\right>_\ell$. For the multi-spot case, we can express the pattern on the DMD and its zero-mean pattern as
\begin{eqnarray}
&&t_\ell(\mathbf{r}) = \Lambda^2 \sum_{m,n} \delta(\mathbf{r} - \mathbf{r}_{mn} - \mathbf{r}_\ell) + t_0 \nonumber \\
&&\Delta t_\ell(\mathbf{r}) \approx  \Lambda^2 \sum_{m,n} \delta(\mathbf{r} - \mathbf{r}_{mn} - \mathbf{r}_\ell),
\label{eqn_multispot_DMD}
\end{eqnarray}
where $\mathbf{r}_{mn} = (m\Lambda, n\Lambda)$, $m$ and $n$ are integers, and $\Lambda$ is the period of the pattern. Then, we can calculate the covariance of the pattern on the DMD as 
\begin{eqnarray}
&&\left<\Delta t_\ell (\mathbf{r}_1) \Delta t_\ell(\mathbf{r}_2)\right>_\ell = \iint \Delta t(\mathbf{r}_1 - \mathbf{r}_\ell) \Delta t(\mathbf{r}_2 - \mathbf{r}_\ell) d^2\mathbf{r}_\ell  \nonumber \\
&&\hspace{0.5in} = \Lambda^4 \sum_{m,n}\delta(\mathbf{r}_1 - \mathbf{r}_2 - \mathbf{r}_{mn}) \star \sum_{m,n}\delta(\mathbf{r}_1 - \mathbf{r}_2 - \mathbf{r}_{mn}) \nonumber \\
&&\hspace{0.5in} \approx \Lambda^4 \eta \sum_{m,n} \delta(\mathbf{r}_1 - \mathbf{r}_2 - \mathbf{r}_{mn}),
\label{eqn_multispot_correlation}
\end{eqnarray}
where $\eta$ is a constant that enforces unit consistency. Plugging this into Eq.~\eqref{eqn_DMD_correlation}, we can then calculate the pattern-pattern covariance as
\begin{eqnarray}
&&\left<\Delta p_\ell(\mathbf{r}) \Delta p_\ell(\mathbf{r}')\right>_\ell = (h_{\mathrm{illu}}\star h_{\mathrm{illu}})(\mathbf{r} - \mathbf{r}') \otimes \Lambda^4 \eta \sum_{m,n} \delta(\mathbf{r} - \mathbf{r}' - \mathbf{r}_{mn}).
\label{eqn_multispot_pattern_correlation}
\end{eqnarray} 
Although the pattern-pattern covariance is only a replica of the $(h_{\mathrm{illu}}\star h_{\mathrm{illu}}) (\mathbf{r})$, the PSF of the covariance image, $I_{\mathrm{cov}}(\mathbf{r})$, only depends on the multiplication of $h_{\mathrm{det}}(\mathbf{r})$ and $(h_{\mathrm{illu}}\star h_{\mathrm{illu}})(\mathbf{r}) \otimes \Lambda^4 \eta \sum_{m,n} \delta(\mathbf{r} - \mathbf{r}_{mn})$ as Eq.~\eqref{eqn_IPE_SIM} derived. If the period of the multi-spot pattern is large compared to $(h_{\mathrm{illu}} \star h_{\mathrm{illu}})(\mathbf{r})$, we can still have our PSF as $[(h_{\mathrm{illu}}\star h_{\mathrm{illu}})\cdot h_{\mathrm{det}}] (\mathbf{r})$, which is the same as the case of random pattern illumination. 

\section*{Appendix B: Enhanced SNR via pixel reassignment} \label{sec_SIMS_PR}

In this section, we first discuss the similarity between SIMS and confocal microscopy. This leads to an extension of our method that incorporates the pixel reassignment procedure proposed in~\cite{Cox1982, Sheppard1988, Muller2010, Roth2013, Sheppard2013}. In computing the covariance of the shifted pattern $p_\ell(\mathbf{r}-\mathbf{r}_s)$ and the intensity $I_\ell(\mathbf{r})$, there is still some information of the object leftover. Pixel reassignment helps incorporate it in a straightforward fashion, giving better SNR in the final reconstruction.

In Sec.~\ref{sec_SIMS} of our SIMS procedure, we first calculate the covariance image $I_{\mathrm{cov}}(\mathbf{r})$. The PSF of this covariance image is determined by imposing our statistical prior on the pattern-pattern covariance $\left< \Delta p_\ell (\mathbf{r}) \Delta p_\ell (\mathbf{r}') \right>_\ell$. The effect is similar to the illumination PSF of confocal microscopy~\cite{Colin2003}. Looking at Eq.~\eqref{eqn_IPE_SIM_DMD}, our covariance image with PSF of $[(h_{\mathrm{illu}}\star h_{\mathrm{illu}})\cdot h_{\mathrm{det}}] (\mathbf{r})$ is the same as a confocal image taken with illumination PSF, $(h_{\mathrm{illu}}\star h_{\mathrm{illu}}) (\mathbf{r})$, and detection PSF, $h_{\mathrm{det}}(\mathbf{r})$. 

From the same SIM dataset, we can further use the shifted patterns $p_\ell(\mathbf{r}-\mathbf{r}_s)$ and correlate them with the intensity $I_\ell(\mathbf{r})$ to compute a series of shifted covariance images
\begin{eqnarray}
&&I_{\mathrm{cov}}^s (\mathbf{r},\mathbf{r}_s) = \left<\Delta p_\ell(\mathbf{r} - \mathbf{r}_s) \Delta I_\ell(\mathbf{r})\right>_\ell = \iint o(\mathbf{r}') \left< \Delta p_\ell (\mathbf{r} - \mathbf{r}_s) \Delta p_\ell(\mathbf{r}')\right>_\ell h(\mathbf{r} - \mathbf{r}') d^2 \mathbf{r}' \nonumber \\
&& \hspace{0.5in} = \iint \alpha_t o(\mathbf{r}') (h_{\mathrm{illu}}\star h_{\mathrm{illu}})(\mathbf{r} - \mathbf{r}_s - \mathbf{r}') h_{\mathrm{det}}(\mathbf{r} - \mathbf{r}') d^2 \mathbf{r}'.
\label{eqn_4D_data}
\end{eqnarray}
The PSF of the shifted covariance image $I_{\mathrm{cov}}^s (\mathbf{r})$ is the product of $(h_{\mathrm{illu}}\star h_{\mathrm{illu}}) (\mathbf{r}-\mathbf{r}_s)$ and $h_{\mathrm{det}}(\mathbf{r})$, whose center is approximately at $\mathbf{r}_s/2$. This image is the same as the image taken under a confocal microscope with a shifted pinhole. This implies by shifting around the patterns and correlating with the intensity, we get the equivalent of many 2D confocal images taken with the pinhole at different positions. This is the same dataset as would be described in the imaging scanning microscope, where the single-pixel camera and pinhole is replaced with a CCD in the confocal system~\cite{Muller2010,Sheppard2013}. Though these images are not centered, they still contain the information of the same object. Pixel reassignment was proposed in~\cite{Sheppard1988, Muller2010, Roth2013, Sheppard2013} as a way to incorporate this 4D information to get a 2D image with better SNR. 

\begin{figure}[h]
\centering
\includegraphics[width=11.4cm]{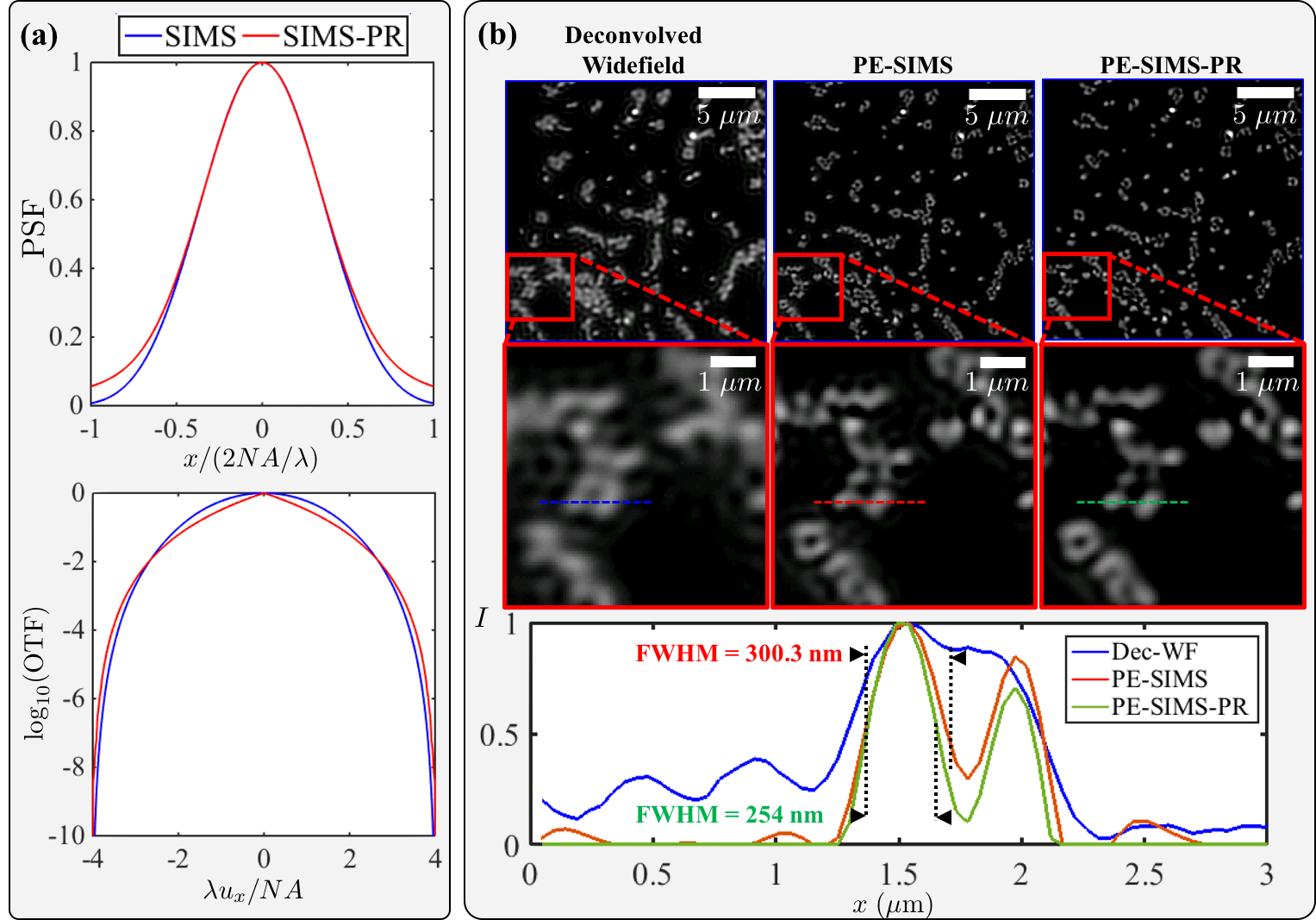}
\caption{(a) Comparison of the PSF and OTF for SIMS and SIMS with pixel reassignment (PR). (b) Comparisons of the deconvolved widefield image and the reconstructions of the $6\times 6$ multi-spot scanned fluorescent beads with and without pixel reassignment.}
\label{fig_SIMS_PR_OTF_PSF}
\end{figure}

Since the 2D images from $\mathbf{r}_s$-shifted patterns are approximately $\mathbf{r}_s/2$-shifted versions of the one at $\mathbf{r}_s = \mathbf{0}$, we can shift the information back to the center region and sum up all these images to enhance the SNR and form a pixel-reassigned (PR) image as 
\begin{eqnarray}
&&I_{\mathrm{PR}} (\mathbf{r}) = \iint I^s_{\mathrm{cov}}\left(\mathbf{r}+ \frac{\mathbf{r}_s}{2}, \mathbf{r}_s\right) d^2 \mathbf{r}_s \nonumber \\
&&\hspace{0.5 in} = \iint \alpha_t o(\mathbf{r}') \left[\iint (h_{\mathrm{illu}}\star h_{\mathrm{illu}})\left(\mathbf{r} - \frac{\mathbf{r}_s}{2} - \mathbf{r}'\right) h_{\mathrm{det}}\left(\mathbf{r} + \frac{\mathbf{r}_s}{2} - \mathbf{r}'\right) d^2 \mathbf{r}_s \right] d^2 \mathbf{r}' \nonumber \\
&&\hspace{0.5 in} =  \iint \alpha_t  o(\mathbf{r}') [(h_{\mathrm{illu}}\star h_{\mathrm{illu}}) \otimes h_{\mathrm{det}}] (2(\mathbf{r} - \mathbf{r}')) d^2 \mathbf{r}' 
\label{eqn_SIMS_PR}
\end{eqnarray}
This synthesized image using pixel reassignment gives a PSF of $[(h_{\mathrm{illu}}\star h_{\mathrm{illu}}) \otimes h_{\mathrm{det}}] (2\mathbf{r})$. Figure~\ref{fig_SIMS_PR_OTF_PSF}(a) shows the comparison between the SIMS PSF, $[(h_{\mathrm{illu}}\star h_{\mathrm{illu}}) \cdot h_{\mathrm{det}}] (\mathbf{r})$, and the PSF of SIMS with pixel reassignment, $[(h_{\mathrm{illu}}\star h_{\mathrm{illu}}) \otimes h_{\mathrm{det}}] (2\mathbf{r})$ both in the real space and the Fourier space (assuming $h_{\mathrm{illu}} \approx h_{\mathrm{det}}$). In the real space, the PSF after doing pixel reassignment looks fatter than the one without pixel reassignment. However, the OTF of the one with pixel reassignment has larger value in the high-frequency region, where the noise severely degrade the image resolution. Thus, we get better SNR by summing up all the information we have and have a OTF that better deals with noise at high-frequency region. Since we know the PSF, $[(h_{\mathrm{illu}}\star h_{\mathrm{illu}}) \otimes h_{\mathrm{det}}] (2\mathbf{r})$, and the shading map, $\alpha_t(\mathbf{r})$, of this pixel-reassigned image $I_{\mathrm{PR}}(\mathbf{r})$, we can again apply the deconvolution and the shading correction operation described in Sec.~\ref{sec_SIMS} to get a PE-SIMS-PR reconstruction.

Figure~\ref{fig_SIMS_PR_OTF_PSF}(b) compares the reconstruction result of fluorescent beads using $6\times 6$ multi-spot illumination with and without applying pixel reassignment algorithm. Pixel reassignment results in sharper contrast when two beads are close to each other and helps clean up some background deconvolution errors. A cut-line plot of the fluorescent beads in Fig.~\ref{fig_SIMS_PR_OTF_PSF}(b) shows that the FWHM of the reconstructed bead from SIMS (300.3 nm) is larger than for SIMS-PR with pixel reassignment (254 nm), giving better resolution.

\end{document}